%% file: iclr2027_conference.tex
\documentclass{article} 
\usepackage{iclr2027_conference,times}
\usepackage[pdftex]{graphicx}
\input{math_commands.tex}

\usepackage{multirow}
\usepackage{hyperref}
\usepackage{url}
\usepackage{subcaption}
\usepackage{capt-of}
\title{From Patching to Pruning Visual Computation in Vision–Language Models}

\author{Rahul Chowdhury$^{1}$ \quad
Timothy A Rupprecht$^{2}$ \quad
Xuan Shen$^{3}$ \quad
Shaoyi Huang$^{4}$ \quad
Pu Zhao$^{1}$ \quad
\\
\textbf{Yanzhi Wang$^{1}$} \quad
\\[2mm]
$^{1}$Northeastern University, USA \\
$^{2}$EmbodyX Inc., USA \\
$^{3}$Zhejiang University, China \\
$^{4}$Stevens Institute of Technology, USA \\
\texttt{\{chowdhury.rah,p.zhao\}@northeastern.edu}
}

\iclrfinalcopy
\begin{document}

\maketitle

\begin{abstract}
Vision--language models (VLMs) incur substantial inference cost because every visual token is processed by the attention and MLP projections of every decoder layer, even when token-specific visual computation is unnecessary at many depths. We introduce \textbf{Patch-to-Prune (P2P)}, inspired by Mechanistic Interpretability, a training-free framework that converts activation patching from a diagnostic tool into an inference-time computation bypass. \textbf{P2P} performs validation-guided forward and backward layer sweeps to identify decoder regions whose visual-token projection outputs can be replaced by fixed neutral proxy activation vectors within a user-specified accuracy tolerance. Unlike conventional token-pruning methods, \textbf{P2P} preserves the sequence length, token order, positional information, attention mask, and residual pathways, thereby pruning computation without removing tokens or modifying the pretrained model weights. We evaluate \textbf{P2P} on four VLMs from the Qwen2.5-VL and LLaVA families across seven multi-modal benchmarks using mutually disjoint calibration, validation, and test partitions. \textbf{P2P} at a 3\% tolerance retains around \textbf{94\%} of dense accuracy while reducing FLOPs by \textbf{55\%}. Beyond these efficiency gains, our layer-wise analysis suggests that visual processing in VLMs is non-uniformly distributed across decoder depth: early and late layers often require little token-specific visual computation, whereas intermediate layers appear to perform most task-relevant visual integration, enabling later reasoning to rely largely on visual information already embedded in shared residual and textual representations. This makes \textbf{P2P} both an efficient inference framework and a causal lens into visual information processing in VLMs.
\end{abstract}

\input{sec/intro}
\input{sec/related}
\input{sec/method2}
\input{sec/evaluation}
\input{sec/conclusion}



\bibliography{iclr2027_conference}
\bibliographystyle{iclr2027_conference}
\newpage
\appendix
\section{Appendix}

\input{apndx/apndxtext}

\end{document}

%% file: math_commands.tex
\usepackage{amsmath,amsfonts,bm}

\def\eqref#1{equation~\ref{#1}}

\def\1{\bm{1}}

\DeclareMathAlphabet{\mathsfit}{\encodingdefault}{\sfdefault}{m}{sl}
\SetMathAlphabet{\mathsfit}{bold}{\encodingdefault}{\sfdefault}{bx}{n}



%% file: sec/intro.tex
\section{Introduction}
Vision--language models (VLMs) have achieved superior performance across visual question answering, multimodal reasoning, object recognition, and spatial understanding. This capability, however, comes with substantial inference cost. An input image is typically represented by hundreds or  thousands of visual tokens, which are 
processed repeatedly by every decoder layer. 
Since visual tokens typically dominate the input sequence, repeatedly applying the attention transformations makes the language decoder a major computational bottleneck in VLM inference \cite{zhao2024fully,shen2025efficient,shen2025fastcar,zhao2025taming,mi2026effective,zhao2026hieramp,li2022pruning,zhang2022advancing,rtseg,yang2023pruning,shen2025quartdepth}.

Prior training-free acceleration methods like FastV ~\cite{chen2024image} and PyramidDrop~\cite{xing2024pyramiddrop} tackle this problem mainly by pruning visual tokens: they estimate each token's importance from early attention or similarity scores, then permanently discard tokens judged less relevant. However, this approach ignores how visual information actually flows through the network, and it overlooks the fact that a token's importance in later layers may not decrease monotonically—a token that seems unimportant early on could become critical downstream.
This limitation raises a complementary question: which layers are actually responsible for processing visual tokens? 



To investigate this problem, we first perform a layer sensitivity analysis through activation patching, which replaces  the output of a selected module with a reference activation  to identify its causal significance. Our   analysis  
reveals a pronounced non-uniformity: substitution of visual-token projection outputs is well tolerated in many early and late layers, whereas an intermediate band of layers exhibits markedly greater sensitivity to such substitution. This pattern suggests that redundancy within VLM decoders is not uniformly distributed, but is instead systematically structured across depth.

\begin{figure}[!t]
\centering
\includegraphics[width=1.0\textwidth]{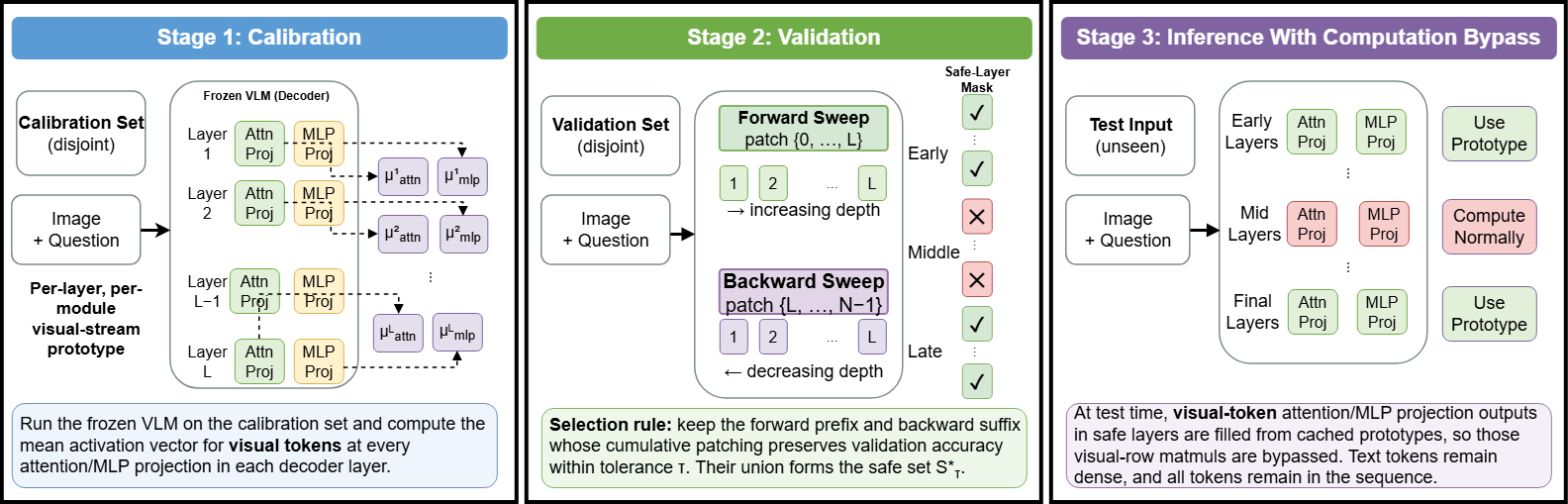}
\caption{Framework of Patch-to-Prune 
}
\label{fig:P2PFW}
\end{figure}
Based on our layer sensitivity analysis,  we propose \textbf{Patch-to-Prune (P2P)} shown in Figure~\ref{fig:P2PFW}, an acceleration   framework that converts this diagnostic intervention into an actual inference-time computation bypass. P2P first runs the frozen VLM on a small calibration set and constructs a modality-conditioned activation prototype for every decoder layer and linear module. It then performs cumulative forward and backward patching sweeps on a disjoint validation set. These sweeps identify the layers  whose visual-token computations can be replaced by   corresponding prototypes while remaining within a specified accuracy tolerance. The union of these  selected regions defines the set of \emph{pressable or safe layers}   during inference.

At test time, P2P does not perform the full computations. Instead, for visual tokens in safe layers, their outputs of attention projection modules 
are  directly filled with  cached activation prototype, without actual computations. Only   textual tokens in safe layers and all tokens in other layers  go through the typical attention computations. 
Importantly, P2P prunes \emph{computation} rather than tokens. 
Input embeddings of visual tokens remain in the sequence throughout the network, and the original sequence length, token order, attention mask, and residual pathways are preserved. This design avoids the irreversible token-removal decisions  by conventional pruning methods and enables P2P to be applied directly to pretrained VLMs without modifying their parameters, changing their architecture, or performing additional training. 
Our main contributions are as follows:
\begin{itemize}
    \item We introduce P2P, an acceleration framework that uses modality-conditioned activation patching to localize redundant visual-token computations across   decoder depth.
    
    \item We present a validation-guided, tolerance-controlled layer-selection procedure that provides an interpretable accuracy--efficiency trade-off. 
    
    \item We convert post-hoc activation replacement into an actual computation bypass that skips selected attention and MLP projection matmuls while preserving all tokens and the original sequence structure. 
    
    \item We demonstrate consistent FLOP and latency reductions across four VLMs and seven multimodal benchmarks, and show that decoder redundancy is concentrated near the early and late layers of the network. We demonstrate that our P2P-aware finetuning can adapt  the model directly under its patched computation graph and substantially recover  the accuracy lost under aggressive pressing.

\end{itemize}

%% file: sec/related.tex
\section{Related Work}


\paragraph{Efficient vision--language model inference.}
VLMs commonly represent an image using hundreds or thousands of visual tokens, which substantially increases the computation cost \cite{zhao2025open,zhao2026omnimem,shen2025draftattention,shen2024search,zhao-etal-2024-pruning,zhan-etal-2024-rethinking-token,shen2024numerical}. To address this,  a  family of methods reduces the visual tokens \cite{kong2025,yang2026survey,zhan2024exploring,shen2025sparse,zhan2024fast,shen2024lazydit}. PruMerge \cite{shang2025prumerge} selects informative visual tokens  and merges redundant tokens into the retained representations. 
PyramidDrop \cite{xing2024pyramiddrop} extends the idea of FastV  \cite{chen2024fastv}  with a progressive schedule that partitions the decoder into stages and removes a fraction of the surviving visual tokens at each stage. 
SparseVLM \cite{zhang2024sparsevlm} uses text-guided attention to rank visual tokens, selects layer-dependent sparsification ratios, and recycles information from discarded tokens. 
Collectively, these methods demonstrate that 
substantial computation can be eliminated.


\paragraph{Activation patching and causal localization.}
Activation patching, also referred to as interchange intervention or causal tracing, determines the causal importance of an internal model component by replacing its activation and measuring the resulting behavioral change \cite{geiger2021causal,zhang2023activationpatching}. It has been used to localize information processing to particular layers, token positions, attention heads, and intermediate representations in language models. Prior work has also studied methodological choices for patching and scalable approximations that reduce the cost of evaluating large numbers of interventions \cite{kramar2024atp,heimersheim2024activation}. These methods are primarily interpretability tools: the original computation is normally executed before its activation is overwritten, so the intervention itself does not  accelerate inference.

%% file: sec/method2.tex
\section{Layer Sensitivity Analysis}
\label{sec:background}

\input{figuretex/motivationpope}

To identify which layers are actually responsible for processing vision and text tokens when the model  answers questions, we first perform a layer sensitivity analysis.

\subsection{Notations}
Let $\mathcal{M}$ be a frozen, instruction-tuned VLM whose language backbone consists of $N$ decoder layers. Given an image--question pair, the processor produces a token sequence of length $T$ that interleaves projected visual tokens with text tokens. Each decoder layer $\ell \in \{0,\dots,N-1\}$ contains a set of linear projections $\mathcal{S} = \{W^Q, W^K, W^V, W^O, W^{\text{gate}}, W^{\text{up}}, W^{\text{down}}\}$, i.e., the four attention projections and the three gated-MLP projections. For a module $m \in \mathcal{S}$ at layer $\ell$, we write $a_{\ell,m}(t) \in \mathbb{R}^{d_m}$ for its output activation at token position $t$, where $d_m$ is the module's output length.

\subsection{Stream-Conditional Activation Statistics}
\label{sec:method:stats}

To identify the layer sensitivity on tokens, we replace the layer outputs with example-level mean activations. Specifically, 
on a statistics split $\mathcal{D}_{\mathrm{stat}}$ of
$N_{\mathrm{stat}}$ training examples (disjoint from all evaluation data), for every layer $\ell$, module $m$, and stream $s \in \{\mathrm{img}, \mathrm{txt}\}$, we run the frozen model and record the example-level mean activation
\begin{equation}
\mu_{\ell,m,s} \;=\; \frac{1}{N_{\mathrm{stat}}} \sum_{n=1}^{N_{\mathrm{stat}}}
\frac{1}{|\mathcal{P}^{(n)}_{s}|} \sum_{t \in \mathcal{P}^{(n)}_{s}}
a^{(n)}_{\ell,m}(t).
\label{eq:mean}
\end{equation}
Note that here each example contributes equally  regardless of its token count, which prevents long sequences (or large images) from dominating the estimate.  

\subsection{Cumulative Layer Intervention}
\label{sec:method:sweeps}

To localize the causally load-bearing depth band, we run
\emph{cumulative} sweeps on a held-out validation split. These sweeps are done on three setups of streams: \texttt{image\_only} (image positions
$\leftarrow$ image means), \texttt{text\_only} (text positions
$\leftarrow$ text means), and \texttt{dual} (both, same-stream).
We perform sweep from two directions with patch layers: 
(i) \textbf{Forward:} patch layers  $\mathcal{L}_{\mathrm{fwd}}(L) = \{0, \dots, L\}$ with $L$ starting from 0 and growing from the input side; and (ii) \textbf{Backward:} patch layers   $\mathcal{L}_{\mathrm{bwd}}(L) = \{L, \dots, N-1\}$ with $L$ starting from $N-1$ and decreasing from  output.
For each value of $L$ during each direction in each stream, we replace the module outputs of patch layers with the  example-level mean activation, and record the corresponding accuracy drop on this split.
Each sweep point re-registers the intervention on the specific layer set and these points localize how much \emph{token-specific} information each modality contributes at a given depth.

\subsection{Observations and Insights}
\label{sec:method:protocol}

\paragraph{Benchmarks.} We evaluate on POPE (tests object-recognition), WhatsUp (tests spatial-relation),  and
ScienceQA-IMG (image-dependent scienctific knowledge). For each dataset, from the training set, we randomly draw  a  statistics split $\mathcal{D}_{\mathrm{stat}}$
(with $N_{\mathrm{stat}} = 100$ examples) used exclusively for
Eq.~\ref{eq:mean}, and another validation split ($100$ examples) to perform sweeps. These splits and the final test set are three mutually disjoint sets.

\paragraph{Models and sweep grid.} We use Qwen2.5-VL-3B-Instruct and
LLaVA-Next (Vicuna-7B and Vicuna-13B). The localization grid crosses
$\{$model$\} \times \{$dataset$\} \times \{$forward, backward$\} \times
\{$\texttt{image\_only}, \texttt{text\_only}, \texttt{dual}$\}$ with mean patching for each value of $L$.

\paragraph{Observations.}
Figure~\ref{fig:preliminary}, Figure~\ref{fig:preliminary_wu}  (Appendix~\ref{app:sweep}) and Figure~\ref{fig:preliminary_sqa} (Appendix~\ref{app:sweep}) 
reveal a consistent depth-dependent pattern: replacing visual-token projection outputs is often tolerated across substantial early and late decoder regions (demonstrated by the small accuracy drop of a few first layers in the forward sweep or a few last layers in the backward sweep), whereas extending the intervention into intermediate layers causes markedly larger accuracy degradation. In contrast, patching text-token computations remains broadly damaging, indicating that this redundancy is primarily associated with the visual stream. We hypothesize that visual information follows a staged computational pathway. Early layers begin encoding visual features, intermediate layers perform the primary task-relevant integration into the shared residual stream and textual representations, and later layers generate the answer largely from information already embedded in these representations. This motivates replacing  unnecessary visual-token projection computations at pressable depths with a modality-conditioned fixed proxy activation. This preserves the average contribution of each layer while bypassing token-specific computation.



\section{Patch-to-Prune Framework}

\label{sec:method}

\subsection{Overview}

We propose \textbf{Patch-to-Prune (P2P)}, a training-free method for reducing the inference cost of VLMs. P2P   framework consists of three stages: (i) layer sensitivity analysis with mechanistic profiling of model's dependence on visual computation at each projection module, (ii) validation-guided safe-layer selection, and (iii) computation-bypassed inference.

\subsection{Layer Sensitivity Analysis}
As discussed in Section~\ref{sec:background}, P2P first performs layer sensitivity analysis with activation patching to identify decoder layers in which visual token computations can be replaced without causing a large decrease in task accuracy. 
It profiles the accuracy drop under various patching layer configurations during forward and backward sweeps.

\subsection{Safe-Layer Selection}

Based on the profiling of the accuracy drop under various patching layer configurations, P2P then selects safe layers. Specifically,  given a tolerance $\tau$,  we  find an admissible patch layer index $L$ such that its cumulative patch
preserves validation accuracy, i.e., $\mathrm{Acc}_d(L) \geq (1-\tau)\,\mathrm{Acc}_0$. We search  $L^{*}_{\mathrm{fwd}}$ and $L^{*}_{\mathrm{bwd}}$ for the forward and backward sweeps respectively. 
In the forward search, we increase $L$ gradually until we find two consecutive \textit{inadmissible} boundaries. Then  the forward frontier is set to $L^{*}_{\mathrm{fwd}} = L-1$. 
The backward search mirrors this from the output side with decreasing $L$. 
The final safe set is the union of the resulting prefix and suffix,
$\mathcal{S}^{*}_{\tau} = \{0,\dots,L^{*}_{\mathrm{fwd}}\} \cup
\{L^{*}_{\mathrm{bwd}},\dots,N{-}1\}$, which are fixed after selection and applied unchanged at test time. The layers  in $\mathcal{S}^{*}_{\tau}$  are safe to remove under a tolerance $\tau$.

\subsubsection{Tolerance Sensitivity.}
\begin{figure*}[t]
    \centering
    \begin{subfigure}[t]{0.48\textwidth}
        \centering
        \includegraphics[width=\linewidth]{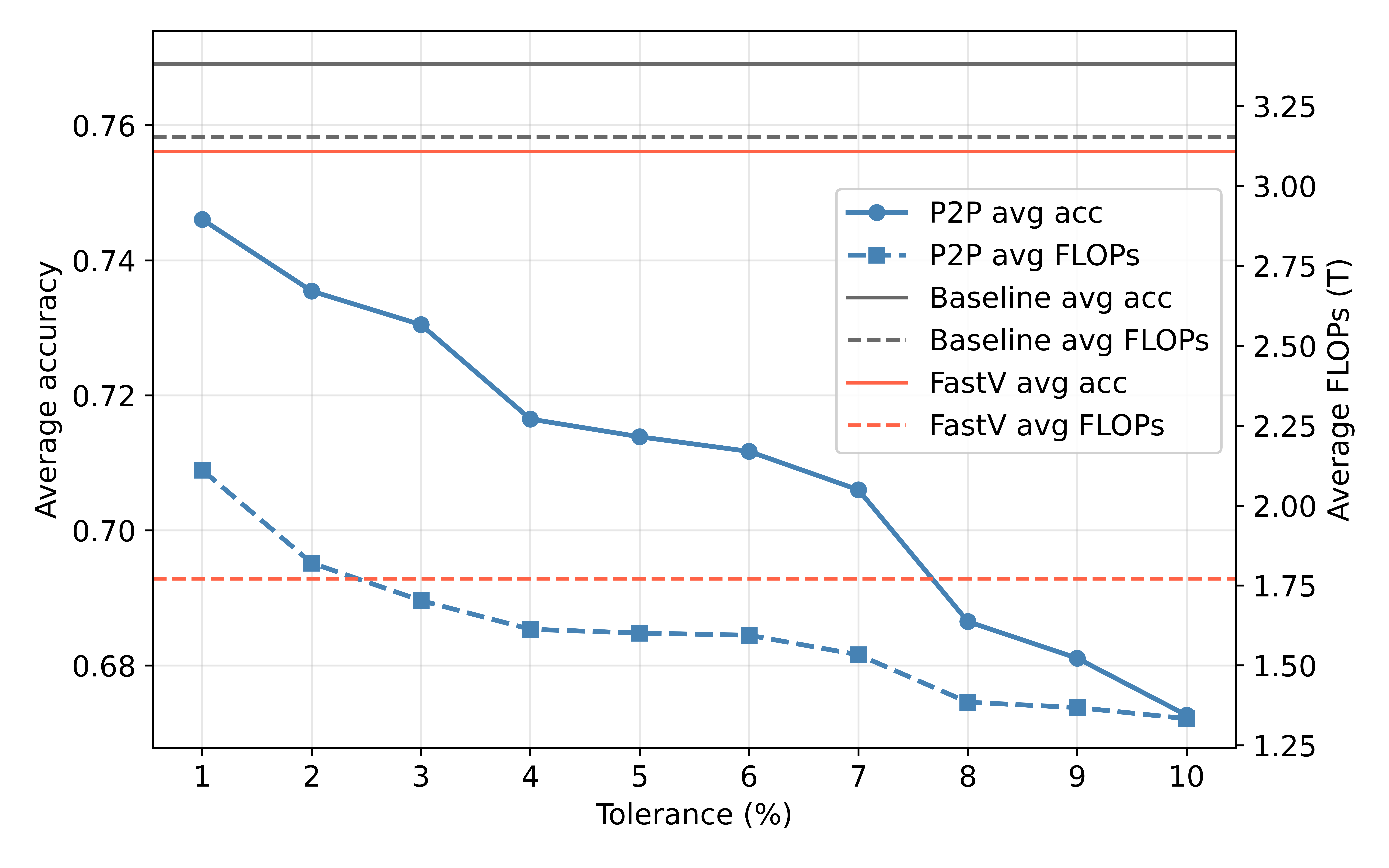}
        \caption{Accuracy and FLOPs as P2P  prunes image-stream computation in more decoder layers under increasing tolerance, compared with Vanilla and FastV.}
        \label{fig:first}
    \end{subfigure}
    \hfill
    \begin{subfigure}[t]{0.48\textwidth}
        \centering
        \includegraphics[width=0.94\linewidth]{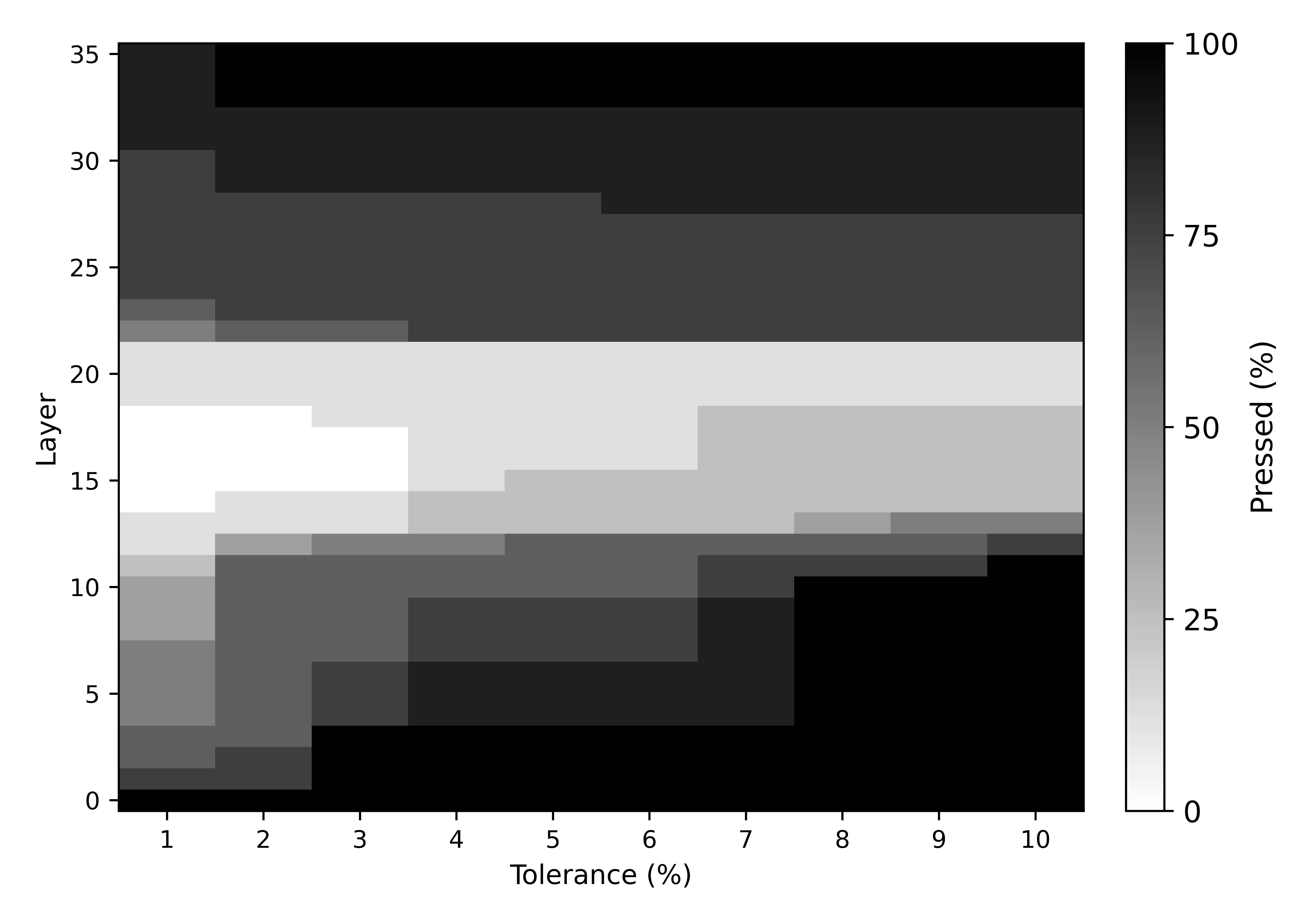}
        \caption{Layer-wise image-stream pruning configurations selected by cumulative forward and backward validation sweeps across datasets and tolerance levels.}
        \label{fig:second}
    \end{subfigure}
    \caption{Tolerance sensitivity analysis.}
    \label{fig:combined}
\end{figure*}
The tolerance $\tau$ is the single control parameter of P2P: relaxing it admits more boundaries under the selection rule above, expanding the safe set inward from both ends of the decoder. To characterize this trade-off on Qwen2.5-VL-3B-Instruct, we sweep $\tau$ from 1\% to 10\%, re-running safe-layer selection on the validation
split at each setting and evaluating the resulting configuration on the full test set. Figure~\ref{fig:combined}(a)  shows that increasing $\tau$ produces a consistent accuracy--efficiency trade-off: average accuracy falls from $74.6\%$ at $1\%$ to $70.6\%$ at $7\%$ and $67.1\%$ at $10\%$,
while average computation drops from $2.1$ to $1.3$ TFLOPs against a $3.1$ TFLOP dense reference. Figure~\ref{fig:combined}(b)   providing the layer-wise view explains this: pressed safe layers first appear at the top of the decoder and near the input,
and as $\tau$ grows they expand inward toward the sensitive intermediate band
(approximately layers 14--21), which is precisely where accuracy begins to break down. We therefore adopt $\tau = 3\%$ as the default operating point with a better trade-off between accuracy and computations. 

\subsection{Computation-Bypassed Inference}
\label{sec:p2p_inference}

Once the pressed set $\mathcal{S}^{*}_{\tau}$ is fixed by safe-layer selection, the diagnostic intervention is converted into an actual computation bypass.
At the selected safe layers,  P2P skips the multiplication with attention and the three MLP modules for visual tokens at  test time. Their module outputs are filled directly
with the cached prototype $\mu^{v}_{l,m}$   discussed in Section~\ref{sec:background}.
Only the remaining textual tokens  go through the original attention computations with the original weights. 


\subsection{P2P-Aware Fine-Tuning}
\label{sec:p2p_finetuning}

We propose an optional P2P-aware finetuning  under  P2P-patched computation graph for accuracy. 

\paragraph{Fine-tuning under mean patching.}
For  the safe layers $\mathcal{S}^{*}_{\tau}$ identified as safe under an accuracy tolerance $\tau$, we insert LoRA adapters   into the attention and MLP projections of these layers. 
To mitigate the training and test mismatch problem, we use the same P2P-patched computation graph during training. Specifically, for visual tokens in safe layers, we skip their computations in projection modules and directly use the cached mean activation in Eq.~\ref{eq:mean} to replace their original outputs. Thus the corresponding projection computation is bypassed  and no gradient is propagated through the LoRA adapter for those patched positions. 
The adapters are instead optimized through the remaining active computation, including unpatched textual tokens  and modules outside the patched layers. This encourages the adapters to compensate for the information removed by P2P rather than relying on activations that are unavailable during patched inference.

\paragraph{Training objective.}
We optimize   adapters using teacher-forced causal language modeling. The multimodal input prompt is masked from the loss, and cross-entropy is computed only over the ground-truth assistant-answer tokens:
\begin{equation}
\mathcal{L}
=
-
\sum_{j\in\mathcal{Y}}
\log
p_{\boldsymbol{\theta}_{0},
\boldsymbol{\theta}_{\mathrm{LoRA}}}^{\mathrm{P2P}}
\left(
y_j
\mid
\mathbf{x},y_{<j}
\right),
\end{equation}
where $\mathcal{Y}$ denotes the  answer-token positions, and $\boldsymbol{\theta}_{\mathrm{LoRA}}$ denotes  LoRA parameters.
Consequently, P2P-aware fine-tuning adapts the model directly to its patched execution  with less training efforts.
We further discuss the distinguishing advantages of P2P in Appendix~\ref{app:advantages}.

%% file: figuretex/motivationpope.tex
\begin{figure*}[!t]
\centering
\includegraphics[width=0.95\textwidth]{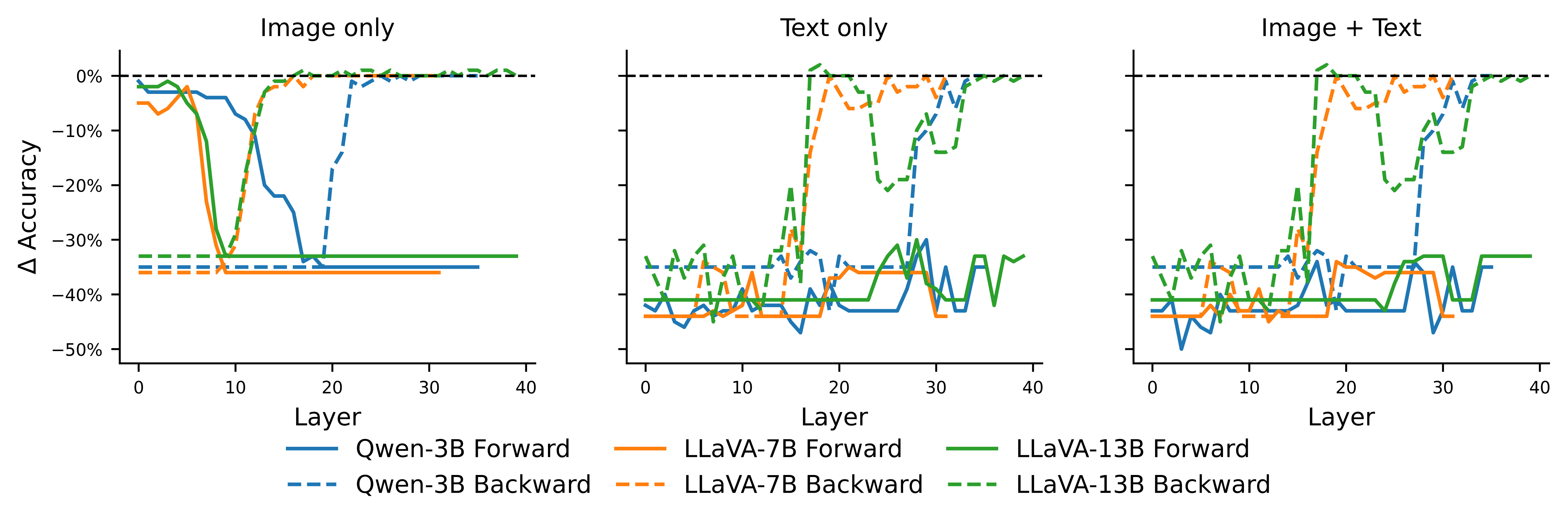} 
\caption{
Layer-wise sensitivity on POPE under cumulative forward and backward mean patching of image, text, and joint image–text token streams.
}
\label{fig:preliminary}
\end{figure*}

%% file: sec/evaluation.tex
\section{Evaluation}

\subsection{Experiment Setup}
\input{sec/baselines}
\input{tables/mainresfinal}
\input{sec/mainresmain}

%% file: sec/baselines.tex
\input{sec/bench}

\paragraph{Models.}
We evaluate P2P on four VLMs from two model families, \textbf{Qwen}~\cite{Qwen-VL, Qwen2VL, qwen2.5-VL} and \textbf{LLaVA}~\cite{liu2023visual,liu2024improved, liu2024llavanext}, spanning parameter scales of 3B, 7B, and 13B. Specifically, we use Qwen2.5-VL-3/7B-Instruct, and LLaVA-v1.6-Vicuna-7/13B-HF.




\paragraph{Baselines.}
\label{sec:baseline_setups}
We compare P2P with the dense  model and two training-free visual-token pruning methods, FastV~\cite{chen2024image} and PyramidDrop~\cite{xing2024pyramiddrop}, implemented within the same evaluation framework using identical models, prompts, precision, test examples, and metrics. The Vanilla baseline applies no token removal, activation replacement, or layer removal. 
We include token pruning baselins (FastV and PyramidDrop) not as competitors but to put P2P in perspective: obviating per-layer visual computation matches the efficiency of token pruning, and opens a complementary direction—pruning tokens \emph{within} the vital layers P2P preserves.

\paragraph{P2P-aware finetuning protocol.}
For main results, we do not finetune the model to make a fair comparison.  In ablation study,  we further demonstrate that our finetuning can improve the performance effectively. 
Our finetuning trains on  the 100-sample statistics split that built the mean prototypes and validates on the 100-sample split that produced the saliency profiles, adding no data beyond the training-free pipeline. 
We train with batch size $1$, LoRA rank $r=1$, scaling factor $\alpha=1$, and AdamW with a learning rate of $2\times10^{-4}$. Training is performed for $10$ epochs. 

%% file: sec/bench.tex
\paragraph{Datasets.}
We evaluate P2P on seven benchmarks spanning diverse multimodal capabilities:\textbf{WhatsUp-VLM} \cite{kamath2023s},\textbf{POPE} (pope)~\cite{li2023evaluating} , \textbf{ScienceQA-IMG} (sqa)~\cite{lu2022learn}, \textbf{GQA} (gqa)~\cite{hudson2019gqa}, \textbf{TextVQA} (tvqa)~\cite{singh2019towards}, \textbf{CV-Bench} (cvb)~\cite{tong2024cambrian}, and \textbf{AI2D} (ai2d)~\cite{kembhavi2016diagram}. For each benchmark, we use 100 samples for proxy estimation and a disjoint set of 100 samples for layer selection. We set the proxy-estimation size to 100 based on the ablation in Figure~\ref{fig:size_pt}. Finally, we evaluate on a large held-out test set. More details are in Appendix~\ref{app:bench}.



%% file: tables/mainresfinal.tex
\begin{table*}[!t]
\centering
\caption{Absolute accuracy, FLOPs (teraFLOPs), and average inference latency (seconds per sample) for LLaVA-7B and 13B across seven benchmarks. We compare Vanilla inference, FastV, PyramidDrop, and P2P at 1\%, 3\%, and 5\% tolerance levels.
}
\scalebox{0.95}{
\begin{tabular}{lllccccccccc}
\hline
Method & Metric & Tol. & wco & wvg & pope & sqa & gqa & tvqa & cvb & ai2d \\
\hline
\multicolumn{12}{c}{\textbf{LLaVA-v1.6-Vicuna-7B}} \\
\hline
Vanilla & Acc. & -- & 0.576 & 0.58 & 0.848 & 0.647 & 0.639 & 0.601 & 0.598 & 0.642 \\
 & FLOPs (T) &  & 32.6 & 30 & 32.5 & 23.6 & 32.2 & 34 & 31.2 & 30.1\\
 & Latency (s) &  & 0.218 & 0.199 & 0.22 & 0.163 & 0.531 & 1.07 & 0.213 & 0.201 \\
\cline{1-12}
FastV & Acc. & -- & 0.559 & 0.561 & 0.842 & 0.646 & \textbf{0.636} & 0.596 & 0.605 & 0.634 \\
 & FLOPs (T) &  & 17.1 & 15.8 & 16.9 & 12.6 & 16.7 & 18 & 16.4 & 15.8 \\
 & Latency (s) &  & 0.241 & 0.214 & 0.235 & 0.164 & 0.561 & 1.16 & 0.231 & 0.213 \\
\cline{1-12}
PDrop & Acc. & -- & \textbf{0.567} & 0.572 & \textbf{0.849} & \textbf{0.65} & 0.632 & \textbf{0.599} & \textbf{0.606} & 0.636\\
 & FLOPs (T) &  & 15.4 & 14.2 & 15.1 & 11.3 & \textbf{15} & \textbf{16.2} & 14.8 & 14.2\\
 & Latency (s) &  & 0.233 & 0.209 & 0.23 & 0.163 & 0.55 & 1.13 & 0.224 & 0.21 \\
\cline{1-12}
P2P & Acc. & 1\% & 0.522 & \textbf{0.635} & 0.848 & 0.646 & 0.63 & 0.534 & 0.556 & 0.64 \\
 & FLOPs (T) &  & 10 & 14.1 & 32.5 & 12.3 & 24 & 20.9 & 9.55 & 16.6\\
 & Latency (s) &  & 0.176 & 0.17 & 0.22 & 0.15 & 0.467 & \textbf{0.924} & 0.166 & \textbf{0.175} \\
\cline{1-12}
P2P & Acc. & 3\% & 0.511 & 0.574 & 0.831 & 0.624 & 0.598 & 0.524 & 0.506 & \textbf{0.641} \\
 & FLOPs (T) &  & \textbf{3.69} & 10.8 & 16 & 7.67 & 21.3 & 19 & 6.96 & 15.8  \\
 & Latency (s) &  & \textbf{0.161} & \textbf{0.168} & 0.184 & 0.146 & \textbf{0.447} & 0.927 & 0.167 & 0.176 \\
\cline{1-12}
P2P & Acc. & 5\% & 0.511 & 0.565 & 0.776 & 0.57 & 0.581 & 0.454 & 0.505 & 0.627 \\
 & FLOPs (T) &  & \textbf{3.69} & \textbf{8.27} & \textbf{14.2} & \textbf{2.38} & 16.7 & 16.2 & \textbf{6.09} & \textbf{13.3}\\
 & Latency (s) &  & 0.164 & 0.179 & \textbf{0.182} & \textbf{0.144} & 0.815 & 1.38 & \textbf{0.163} & 0.281 \\
\hline
\multicolumn{12}{c}{\textbf{LLaVA-v1.6-Vicuna-13B}} \\
\hline
Vanilla & Acc. & -- & 0.754 & 0.744 & 0.821 & 0.675 & 0.646 & 0.622 & 0.605 & 0.676 \\
 & FLOPs (T) &  & 62.7 & 57.7 & 62.5 & 45.7 & 61.9 & 65.5 & 60.1 & 58 \\
 & Latency (s) &  & 0.383 & 0.35 & 0.386 & 0.289 & 0.94 & 1.88 & 0.372 & 0.357\\
\cline{1-12}
FastV & Acc. & -- & 0.736 & 0.739 & 0.807 & 0.669 & 0.641 & 0.615 & 0.603 & 0.675  \\
 & FLOPs (T) &  & 32.8 & 30.3 & 32.3 & 24.2 & 32 & 34.4 & 31.5 & 30.4 \\
 & Latency (s) &  & 0.367 & 0.326 & 0.358 & 0.255 & 0.864 & 1.78 & 0.353 & 0.329 \\
\cline{1-12}
PDrop & Acc. & -- & 0.735 & 0.736 & \textbf{0.819} & 0.67 & \textbf{0.646} & \textbf{0.621} & \textbf{0.606} & \textbf{0.677}  \\
 & FLOPs (T) &  & 29.8 & 27.5 & 29.3 & 22 & \textbf{28.9} & \textbf{31.3} & 28.6 & 27.6 \\
 & Latency (s) &  & 0.361 & 0.327 & 0.352 & 0.253 & 0.855 & 1.76 & 0.348 & 0.328 \\
\cline{1-12}
P2P & Acc. & 1\% & \textbf{0.762} & \textbf{0.769} & 0.815 & \textbf{0.674} & 0.594 & 0.613 & 0.537 & 0.676 \\
 & FLOPs (T) &  & 49.9 & 51 & 28.2 & 22.7 & 32 & 62.5 & 14 & 58 \\
 & Latency (s) &  & 0.344 & 0.331 & 0.291 & 0.239 & 0.698 & 1.79 & 0.248 & 0.357 \\
\cline{1-12}
P2P & Acc. & 3\% & 0.726 & 0.748 & 0.811 & 0.661 & 0.594 & 0.564 & 0.537 & 0.613  \\
 & FLOPs (T) &  & 18.8 & 19.8 & 18.2 & 18.6 & 32 & 39 & 14 & 18.5 \\
 & Latency (s) &  & 0.27 & 0.257 & \textbf{0.266} & \textbf{0.227} & 0.698 & 1.48 & 0.247 & 0.255 & \\
\cline{1-12}
P2P & Acc. & 5\% & 0.675 & 0.664 & 0.758 & 0.651 & 0.55 & 0.56 & 0.392 & 0.588  \\
 & FLOPs (T) &  & \textbf{16} & \textbf{17.2} & \textbf{15.3} & \textbf{13.4} & 29.2 & 36.1 & \textbf{7.21} & \textbf{15.9} \\
 & Latency (s) &  & \textbf{0.261} & \textbf{0.255} & 0.323 & 2.11 & \textbf{0.681} & \textbf{1.44} & \textbf{0.235} & \textbf{0.248}\\
\hline
\end{tabular}}
\label{tab:llava7b_13b_acc_flops_latency_absolute}
\end{table*}

%% file: sec/mainresmain.tex
\subsection{Main Result for P2P}
\label{sec:main_results}
\paragraph{From mechanistic discovery to full-test acceleration.}
P2P tests whether visual-computation redundancy discovered from small,
disjoint subsets generalizes to the full held-out test set. 
We evaluate three operating points: a conservative $1\%$ tolerance, the
default $3\%$ tolerance, and a more aggressive $5\%$ tolerance, which
progressively expand the pressed regions to trade accuracy for greater
computation reduction.

\paragraph{Accuracy--computation trade-off.}
Table~\ref{tab:llava7b_13b_acc_flops_latency_absolute} shows that at the conservative $1\%$ tolerance, P2P preserves most of the Vanilla
model accuracy while already removing a substantial amount of
computation. For LLaVA-7B, average accuracy (computed from Table~\ref{tab:llava7b_13b_acc_flops_latency_absolute}) decreases only from
$0.641$ to $0.626$, retaining $97.7\%$ of Vanilla accuracy, while average
FLOPs decrease from $30.8$ to $17.5$~T, corresponding to a
$43.2\%$ reduction. For LLaVA-13B, P2P retains $98.1\%$ of Vanilla
accuracy, with average accuracy changing from $0.693$ to $0.680$, while
reducing FLOPs from $59.2$ to $39.8$~T, a $32.8\%$ reduction. The default $3\%$ tolerance provides a stronger
accuracy--efficiency trade-off. Table~\ref{tab:llava7b_13b_acc_flops_latency_absolute} shows that on LLaVA-7B, P2P reduces average FLOPs
from $30.8$ to $12.6$~T, a $59.1\%$ reduction, while retaining
$93.8\%$ of Vanilla average accuracy. On LLaVA-13B, average FLOPs decrease from $59.2$ to $22.4$~T, a $62.2\%$ reduction, while retaining
$94.8\%$ of Vanilla accuracy. Averaged across the two model sizes, P2P at
$3\%$ therefore retains approximately $94\%$ of Vanilla accuracy while
reducing the reported FLOPs by approximately $60\%$. Increasing the tolerance to $5\%$ further reduces average FLOPs to
$10.1$~T for LLaVA-7B and $18.8$~T for LLaVA-13B, corresponding to
reductions of $67.2\%$ and $68.2\%$, respectively. However, the average
accuracies decrease to $0.574$ and $0.605$. The tolerance therefore
controls a genuine trade-off: relaxing it expands the pressed prefix
and suffix toward   task-sensitive intermediate layers, yielding
greater computation reduction but also increasing the risk of removing task-relevant visual processing.
More results on Qwen models are in Table~3 of Appendix~\ref{app:qwen}.

\paragraph{Comparison with the baseline pruning.}

At the default $3\%$ setting, P2P reaches a substantially lower
computation regime than FastV and PyramidDrop. For LLaVA-7B, P2P
requires $12.6$~T FLOPs on average, compared with $16.2$~T for FastV
and $14.5$~T for PyramidDrop. For LLaVA-13B, P2P requires $22.4$~T,
compared with $31.0$~T for FastV and $28.1$~T for PyramidDrop. P2P
therefore reduces average FLOPs by an additional $22.2\%$ relative to
FastV and $13.1\%$ relative to PyramidDrop on LLaVA-7B, and by
$27.7\%$ and $20.3\%$, respectively, on LLaVA-13B.

This more aggressive computation reduction is accompanied by lower
average accuracy than the token-pruning baselines. At $3\%$, P2P
obtains average accuracies of $0.601$ and $0.657$ on LLaVA-7B and
LLaVA-13B, whereas FastV obtains $0.635$ and $0.686$, and PyramidDrop
obtains $0.639$ and $0.689$. Thus, P2P provides a distinct operating point rather than uniformly dominating baselines. 


\paragraph{Wall-clock acceleration.}
The reduction in visual-token projection operations translates into
measurable wall-clock improvements at the default operating point. In Table~\ref{tab:llava7b_13b_acc_flops_latency_absolute}, on LLaVA-7B, P2P at $3\%$ reduces average latency by $15.4\%$: from $0.351$ to
$0.297$ seconds per sample, whereas FastV and PyramidDrop show higher latency than Vanilla despite their FLOP reductions. On LLaVA-13B,
average latency decreases from $0.619s$ to $0.462s$, a
$25.4\%$ reduction. P2P is also faster than both token-pruning
baselines at this setting: FastV and PyramidDrop require $0.377s$ and
$0.369s$  on LLaVA-7B and $0.578s$ and $0.573s$ seconds on
LLaVA-13B, respectively.

These results are notable since lower theoretical FLOPs need not translate into lower latency. Token-pruning baselines add ranking, indexing, and sequence-reorganization overhead, whereas P2P directly skips selected visual-row projection matmuls. Yet P2P is not strictly monotonic either: because each patched layer carries a fixed patching cost, a higher tolerance can reduce FLOPs while increasing latency, so the lowest-FLOP configuration is not always the fastest.
Overall, the $3\%$
configuration therefore provides the strongest overall balance among
accuracy, FLOPs, and observed wall-clock latency in
Table~\ref{tab:llava7b_13b_acc_flops_latency_absolute}.

\paragraph{Generalization across tasks.}
The full-test results also show that the amount of pressable visual
computation is task dependent. At $1\%$, P2P improves
LLaVA-7B accuracy on WhatsUp-VGQA from $0.580$ to $0.635$. For
LLaVA-13B, it improves WhatsUp-COCOQA from $0.754$ to $0.762$ and
WhatsUp-VGQA from $0.744$ to $0.769$. At $3\%$, LLaVA-7B nearly
preserves its dense AI2D accuracy, obtaining $0.641$ compared with
$0.642$. 
Other tasks, particularly TextVQA and CV-Bench, do not collapse but are more sensitive to aggressive pressing. We note that the full-test accuracy drop can, on some datasets, exceed the specified tolerance. This motivates a fine-tuning setup that tests whether lightweight LoRA adaptation can compensate for that sensitivity. 

Overall, despite this dataset-specific variation, the pressable regions identified yield on an average over all four models and 8 datasets a 55\% FLOPs reduction and 22\% latency reduction while retaining 94\% of dense accuracy—consistent computation savings without collapse, confirming that lightweight mechanistic discovery and activation replacements  generalize to large-scale evaluation.



\input{sec/ablate_imp}

\subsection{Ablation Study} \label{sec:ablation}

\paragraph{Proxy-estimation choice.}
We use example-level mean activation as the proxy estimation.
To justify this choice, we perform ablation by using different replacements: the zero vector (\texttt{zero}); Gaussian noise with per-dimension scale derived from $\sigma_{\ell,m,s}$ or a fixed override
 (\texttt{noise}); or a single real token activation drawn uniformly from  the cached bank (\texttt{random\_token}).
The results in Figure~\ref{fig:ablation_pt} show that our mean activation leads to the best performance.

\paragraph{Proxy-estimation set size ablation.}
We further study how many examples are required to estimate the proxy activations. We vary the statistics-set size over
$N_{\mathrm{stat}}\in\{1,25,50,100,250,500\}$ while keeping the 100-example validation sweep and all other settings unchanged. Our results in Figure~\ref{fig:size_pt} show that 100 statistics examples  provide a stable proxy efficiently. 

\paragraph{Accuracy recovery with P2P-aware fine-tuning.}
\input{tables/finetuning}

Table~\ref{tab:ftp2p} compares the frozen dense model, denoted
\emph{Vanilla}, with frozen P2P and P2P-aware LoRA. P2P-aware fine-tuning
consistently improves over the corresponding frozen P2P configuration
across ScienceQA, TextVQA, and CV-Bench while preserving the same
patched inference graph. At $1\%$,  P2P-aware LoRA
recovers approximately $75\%$ of the average accuracy gap introduced by
pressing, \textbf{$3\%$} recovers approximately \textbf{$100\%$}, and $5\%$ recovers approximately $59\%$. The strong recovery makes $3\%$ the default  tolerance.
P2P-aware LoRA improves accuracy of ScienceQA from $0.624$ to $0.629$, TextVQA
from $0.524$ to $0.589$, and CV-Bench from $0.506$ to $0.628$. P2P shows that diagnostic proxy activations serve as practical substitutes for visual-token computation—effective training-free, and stronger with LoRA adaptation.

%% file: sec/ablate_imp.tex


\begin{figure*}[!t]
\centering
\includegraphics[width=\textwidth]{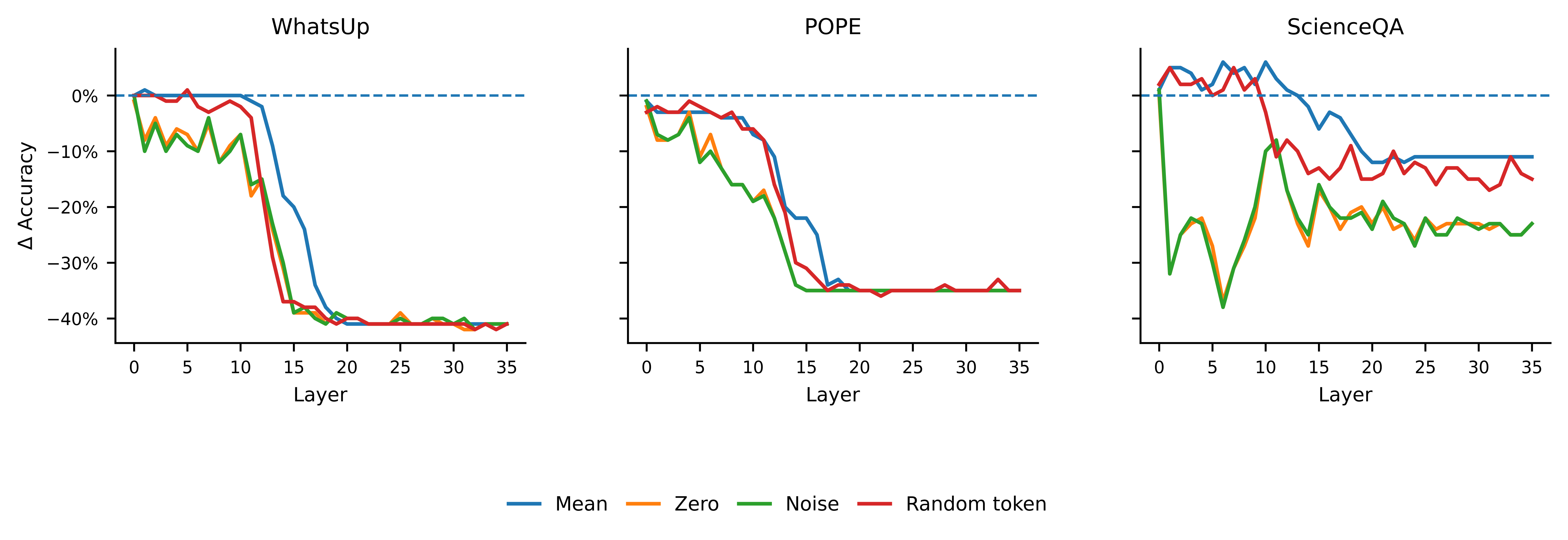} 
\caption{
Patch-type ablation using cumulative forward pruning of image-stream attention and MLP computations with mean, zero, noise, or random-token activation replacement.
}
\label{fig:ablation_pt}
\end{figure*}

\begin{figure*}[!t]
\centering
\includegraphics[width=\textwidth]{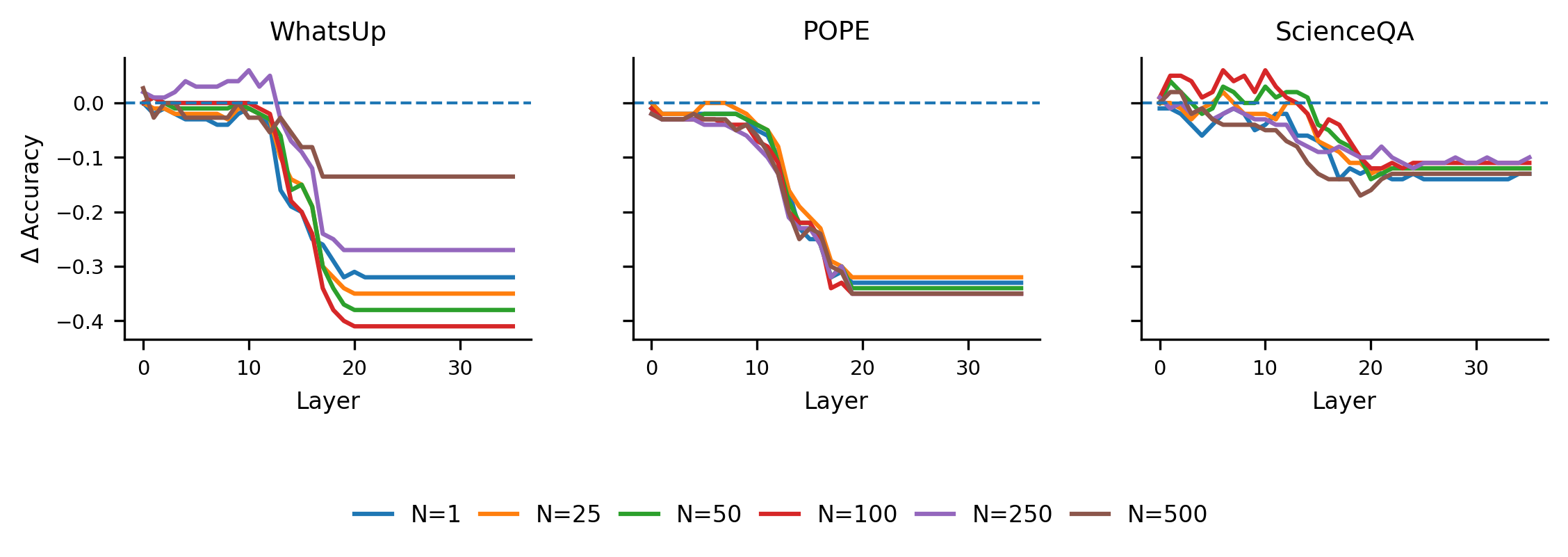} 
\caption{
Proxy-estimation set-size ablation using cumulative forward mean patching of image-stream attention and MLP computations with 1, 25, 50, 100, 250, or 500 statistics examples.
}
\label{fig:size_pt}
\end{figure*}

%% file: tables/finetuning.tex
\begin{table}[!t]

\centering
\caption{Full-test accuracy for LLaVA-7B comparing frozen Vanilla, frozen P2P, and P2P+LoRA fine-tuned under mean patching at 1\%, 3\%, and 5\% tolerances}

\begin{tabular}{llccc}
\hline
  Method & Tol. & sqa & tvqa & cvb \\
\hline
 Vanilla & -- & 0.647 & 0.601 & 0.598 \\
\hline
  P2P & 1\% & 0.646 & 0.543 & 0.556 \\
  P2P+LoRA & 1\% & 0.665 & 0.59 & 0.566 \\
\hline
  P2P & 3\% & 0.624 & 0.524 & 0.506 \\
  P2P+LoRA & 3\% & 0.629 & 0.589 & 0.628 \\
\hline
  P2P & 5\% & 0.570 & 0.454 & 0.505 \\
  P2P+LoRA & 5\% & 0.584 & 0.544 & 0.587 \\
\hline

\end{tabular}
\label{tab:ftp2p}
\end{table}

%% file: sec/conclusion.tex
\section{Conclusion}

P2P turns activation patching from a diagnostic tool into a practical inference mechanism: by identifying decoder depths where visual-token projection outputs can be safely replaced with fixed proxies, it bypasses unnecessary computation without removing tokens or disrupting the original sequence structure. Beyond its efficiency gains, P2P reveals that visual processing is concentrated in a task-sensitive intermediate region, while substantial early- and late-layer computation is redundant. This depth-aware perspective opens a promising direction for token pruning, where future methods can  decide not only \emph{which} visual tokens matter, but also \emph{where} their computation is truly necessary. 

%% file: apndx/apndxtext.tex
\subsection{Advantages}

\label{app:advantages}
\paragraph{Bi-directional cumulative sweeps with unique observations and insights.}
Our  sensitivity analysis show that replacing visual-token projection outputs is often tolerated across substantial early and late layers, while  replacing textual token computations remains  damaging. Our unique observations provide deep insights for how visual information flows in VLMs, as detailed in Section~\ref{sec:method:protocol}.

\paragraph{Pruning computations instead of tokens.}
P2P prunes computations along depth without modifying tokens, whereas prior methods prune tokens along the sequence. Our method is orthogonal and complementary to exist token pruning research.

\paragraph{Finetuning under P2P framework.}
Different from other methods finetuning the whole model with LoRA, our  P2P-aware fine-tuning places  LoRA adapters only in safe layers and keeps patches active during training.
The adapters are consequently optimized under the same computation graph used at inference and learn to compensate for the information removed by the intervention. 


\paragraph{Minor memory cost for caches.}

P2P introduces only a minor memory overhead as it stores a single activation vector for each projection module at each decoder layer. These fixed vectors are reused across all visual tokens, making the cache independent of sequence length and batch size. Thus, P2P trades a small amount of static memory for substantially reduced visual-token computation.
\input{apndx/tabs/other_res_patched}
\input{apndx/fig}

\subsection{Sweep Results}
\label{app:sweep}
We demonstrate the sweep results for WhatsUp in Figure~\ref{fig:preliminary_wu} and   ScienceQA  in Figure~\ref{fig:preliminary_sqa}.

\subsection{Main Results for Qwen Models}
\label{app:qwen}
We demonstrate the Qwen results in Table~\ref{tab:all_models_baselines_p2p_acc_flops}.

\subsection{Patch-Type Ablation}
\label{app:patch_type_ablation}

We examine whether the observed layer sensitivity depends on the
modality-conditioned mean proxy or can be reproduced by arbitrary
activation replacement. Using Qwen2.5-VL-3B-Instruct, we perform the
same cumulative forward sweep on WhatsUp, POPE, and ScienceQA while
replacing the image-token outputs of the attention and MLP projections
with one of four alternatives: the image-stream mean, zero, Gaussian
noise, or a randomly sampled real visual-token activation. As shown in
Figure~\ref{fig:ablation_pt}, mean replacement most consistently
preserves the dense behavior across the pressable early layers. Zero
and noise introduce larger immediate degradation, particularly on
ScienceQA, showing that simply suppressing or perturbing the visual
contribution does not provide a suitable computation proxy. Random
token replacement can remain competitive in some layers because it
retains a realistic activation scale, but its behavior is less stable
and depends on the sampled activation. These results support the
modality-conditioned mean as a deterministic, distributionally
grounded proxy that preserves the typical contribution of each module
while removing example-specific visual computation.

\subsection{Proxy-Estimation Set Size Ablation}
\label{app:proxy_size_ablation}

We further study how many examples are required to estimate the proxy
activations. We vary the statistics-set size over
$N_{\mathrm{stat}}\in\{1,25,50,100,250,500\}$ while keeping the
100-example validation sweep and all other settings unchanged.
Figure~\ref{fig:size_pt} shows that the overall
depth-dependent sensitivity pattern remains similar across all sizes:
the early layers tolerate mean replacement, whereas extending the
intervention toward the intermediate layers causes a pronounced
accuracy drop. Larger statistics sets do not yield a consistent
improvement across WhatsUp, POPE, and ScienceQA, and the transition
into the sensitive depth region is already visible with relatively
few examples. We therefore use 100 statistics examples throughout the
main experiments, as this provides a stable layer- and module-specific
proxy while keeping the mechanistic discovery stage inexpensive.
\subsection{Benchmarks}
\label{app:bench}
\textbf{WhatsUp-VLM} \cite{kamath2023s} evaluates controlled spatial-relation understanding through its COCOQA (wco) and VGQA (wvg) subsets. 1,251 samples were used for testing. \textbf{POPE} (pope)~\cite{li2023evaluating} measures object hallucination using binary object-presence questions, and 8,800 samples were used for testing. \textbf{ScienceQA-IMG} (sqa)~\cite{lu2022learn} evaluates image-grounded scientific reasoning, and all 2,017 image-containing samples were used for testing. \textbf{GQA} (gqa)~\cite{hudson2019gqa} measures compositional reasoning over objects, attributes, and relations, with 12,378 samples for testing. \textbf{TextVQA} (tvqa)~\cite{singh2019towards} evaluates reading and reasoning over scene text, and the complete 5,000 samples were used for testing. \textbf{CV-Bench} (cvb)~\cite{tong2024cambrian} evaluates 2D and 3D vision-centric understanding and 2,438 samples were used for testing. Finally, \textbf{AI2D} (ai2d)~\cite{kembhavi2016diagram} evaluates reasoning over scientific diagrams, and 2,888 AI2D samples were used for testing.

%% file: apndx/tabs/other_res_patched.tex
\begin{table*}[t]
\centering
\caption{Full-test accuracy for LLaVA-7B comparing frozen Vanilla, frozen P2P, and P2P+LoRA fine-tuned under mean patching at 1\%, 3\%, and 5\% tolerances.}

\begin{tabular}{clllcccccccc}
\hline
 & Method & Metric & Tol. & wco & wvg & pope & sqa & gqa & tvqa & cvb & ai2d \\
\hline

\multirow{12}{*}{\rotatebox{90}{\textbf{Qwen2.5-VL-3B-Instruct}}} & Vanilla & Acc. & -- & 0.848 & 0.839 & 0.875 & 0.752 & 0.6 & 0.773 & 0.696 & 0.771 \\
 & Vanilla & FLOPs (T) & -- & 2.39 & 1.93 & 2.27 & 1.6 & 2.22 & 6.31 & 5.71 & 3.16 \\
\cline{2-12}
 & FastV & Acc. & -- & 0.828 & 0.808 & \textbf{0.875} & \textbf{0.752} & \textbf{0.582} & 0.765 & \textbf{0.687} & 0.753 \\
 & FastV & FLOPs (T) & -- & 1.46 & 1.22 & 1.32 & 1.02 & 1.28 & 3.46 & 2.59 & 1.82 \\
\cline{2-12}
 & PDrop & Acc. & -- & 0.801 & 0.77 & 0.848 & 0.735 & 0.568 & 0.737 & 0.663 & 0.717 \\
 & PDrop & FLOPs (T) & -- & 1.34 & 1.13 & 1.2 & 0.954 & 1.16 & \textbf{3.14} & 2.37 & 1.67 \\
\cline{2-12}
 & P2P & Acc. & 1\% & 0.85 & \textbf{0.847} & 0.862 & 0.653 & 0.542 & \textbf{0.767} & 0.678 & \textbf{0.768} \\
 & P2P & FLOPs (T) & 1\% & 1.06 & 0.87 & 1.46 & 0.576 & 1.52 & 6.12 & 3.21 & 2.08 \\
\cline{2-12}
 & P2P & Acc. & 3\% & \textbf{0.85} & \textbf{0.847} & 0.858 & 0.648 & 0.517 & 0.725 & 0.639 & 0.76 \\
 & P2P & FLOPs (T) & 3\% & \textbf{0.958} & 0.87 & 1.13 & 0.511 & 1.14 & 5.21 & 2.02 & 1.79 \\
\cline{2-12}
 & P2P & Acc. & 5\% & \textbf{0.85} & 0.831 & 0.526 & 0.642 & 0.316 & 0.182 & 0.639 & 0.733 \\
 & P2P & FLOPs (T) & 5\% & \textbf{0.958} & \textbf{0.829} & \textbf{0.968} & \textbf{0.412} & \textbf{0.976} & 5.21 & \textbf{1.89} & \textbf{1.57} \\
\hline
\multirow{12}{*}{\rotatebox{90}{\textbf{Qwen2.5-VL-7B-Instruct}}} & Vanilla & Acc. & -- & 0.913 & 0.917 & 0.833 & 0.83 & 0.607 & 0.717 & 0.755 & 0.819 \\
 & Vanilla & FLOPs (T) & -- & 5.58 & 4.5 & 5.3 & 3.74 & 5.18 & 14.5 & 12.9 & 7.28 \\
\cline{2-12}
 & FastV & Acc. & -- & 0.754 & 0.723 & \textbf{0.819} & 0.787 & \textbf{0.563} & 0.647 & 0.742 & 0.773 \\
 & FastV & FLOPs (T) & -- & 3.45 & 2.87 & 3.12 & 2.42 & 3.03 & 8.14 & 6.07 & 4.29 \\
\cline{2-12}
 & PDrop & Acc. & -- & 0.729 & 0.688 & 0.774 & 0.763 & 0.545 & 0.636 & 0.707 & 0.727 \\
 & PDrop & FLOPs (T) & -- & 3.14 & 2.64 & 2.81 & 2.23 & 2.72 & \textbf{7.29} & 5.46 & 3.88 \\
\cline{2-12}
 & P2P & Acc. & 1\% & \textbf{0.891} & \textbf{0.903} & 0.833 & \textbf{0.806} & 0.533 & \textbf{0.71} & \textbf{0.757} & \textbf{0.792} \\
 & P2P & FLOPs (T) & 1\% & 2.68 & 2.16 & 5.3 & 2.63 & 2.4 & 13.5 & 9.37 & 4.11 \\
\cline{2-12}
 & P2P & Acc. & 3\% & 0.89 & 0.863 & 0.81 & 0.804 & 0.517 & 0.639 & 0.734 & 0.769 \\
 & P2P & FLOPs (T) & 3\% & 2.52 & \textbf{1.79} & 2.34 & 2.14 & 2.23 & 12.1 & 6.18 & 3.68 \\
\cline{2-12}
 & P2P & Acc. & 5\% & 0.885 & 0.863 & 0.788 & 0.766 & 0.475 & 0.638 & 0.641 & 0.704 \\
 & P2P & FLOPs (T) & 5\% & \textbf{2.36} & \textbf{1.79} & \textbf{2.17} & \textbf{1.84} & \textbf{2.07} & 11.7 & \textbf{3.78} & \textbf{3.02} \\
\hline
\end{tabular}
\label{tab:all_models_baselines_p2p_acc_flops}
\end{table*}

%% file: apndx/fig.tex
\begin{figure*}[!t]
\centering
\includegraphics[width=\textwidth]{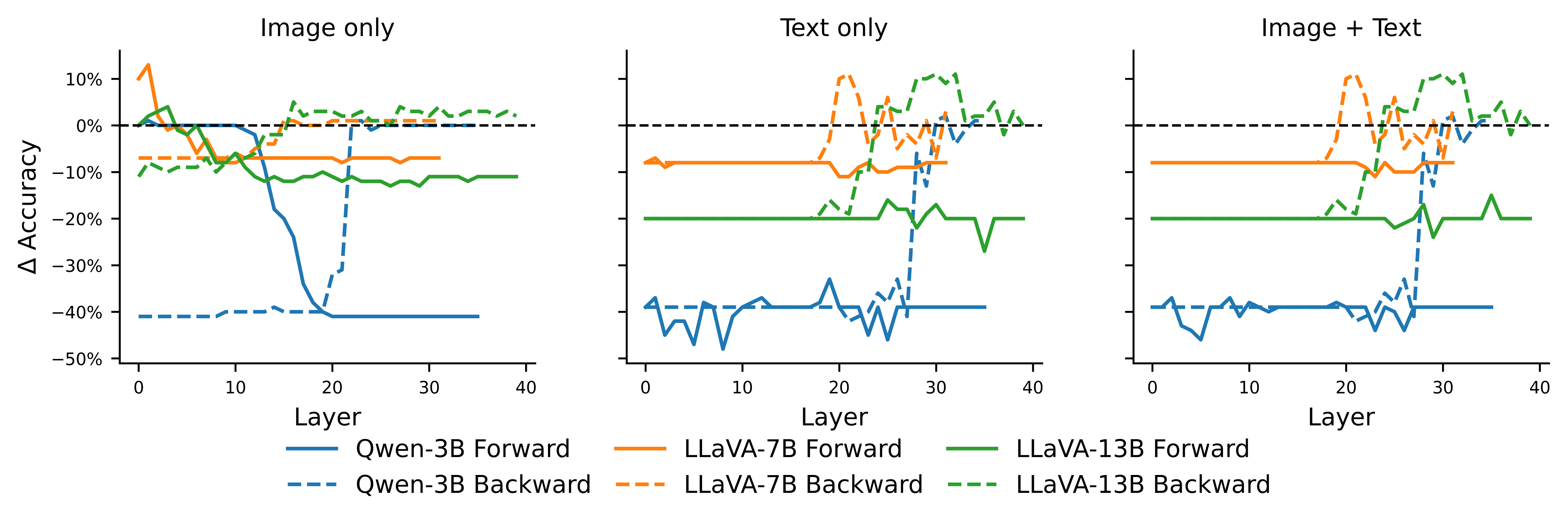} 
\caption{
Layer-wise sensitivity on WhatsUp's COCOQA under cumulative forward and backward mean patching of image, text, and joint image–text token streams.
}
\label{fig:preliminary_wu}
\end{figure*}
\begin{figure*}[!t]
\centering
\includegraphics[width=\textwidth]{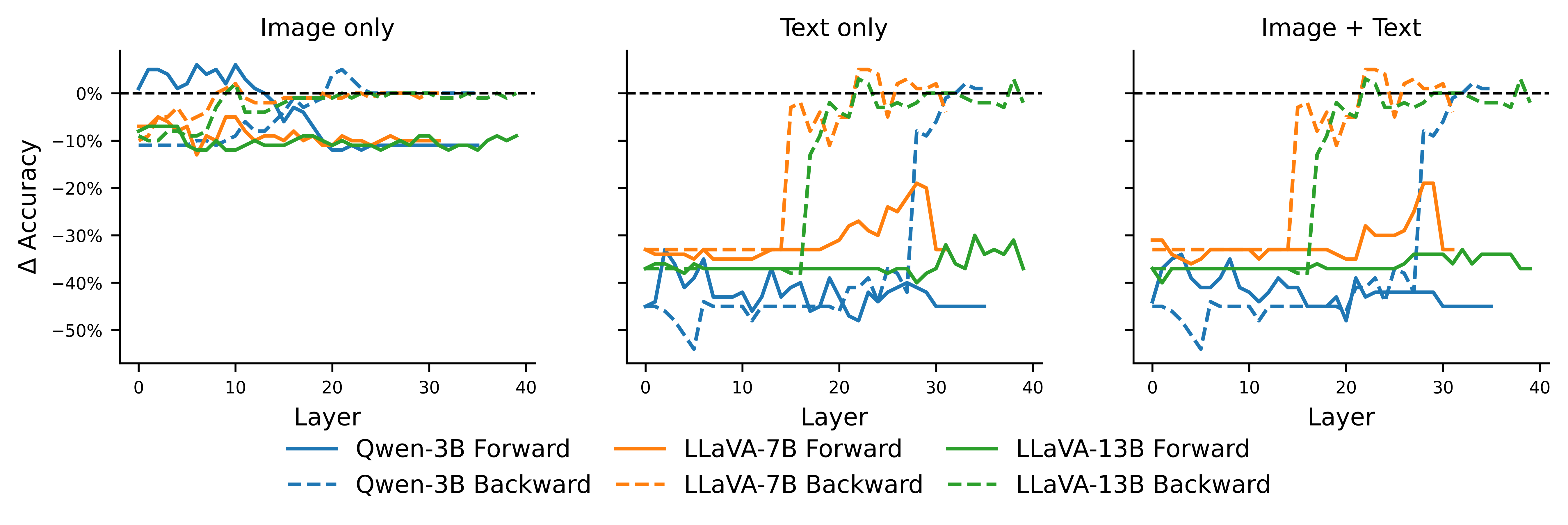} 
\caption{
Layer-wise sensitivity on ScienceQA under cumulative forward and backward mean patching of image, text, and joint image–text token streams.
}
\label{fig:preliminary_sqa}
\end{figure*}

%% file: iclr2027_conference.bib
@inproceedings{kamath2023s,
  title={What’s “up” with vision-language models? investigating their struggle with spatial reasoning},
  author={Kamath, Amita and Hessel, Jack and Chang, Kai-Wei},
  booktitle={Proceedings of the 2023 Conference on Empirical Methods in Natural Language Processing},
  pages={9161--9175},
  year={2023}
}

@inproceedings{li2023evaluating,
  title={Evaluating object hallucination in large vision-language models},
  author={Li, Yifan and Du, Yifan and Zhou, Kun and Wang, Jinpeng and Zhao, Xin and Wen, Ji-Rong},
  booktitle={Proceedings of the 2023 conference on empirical methods in natural language processing},
  pages={292--305},
  year={2023}
}

@inproceedings{lu2022learn,
    title={Learn to Explain: Multimodal Reasoning via Thought Chains for Science Question Answering},
    author={Lu, Pan and Mishra, Swaroop and Xia, Tony and Qiu, Liang and Chang, Kai-Wei and Zhu, Song-Chun and Tafjord, Oyvind and Clark, Peter and Ashwin Kalyan},
    booktitle={The 36th Conference on Neural Information Processing Systems (NeurIPS)},
    year={2022}
}

@inproceedings{hudson2019gqa,
  title={Gqa: A new dataset for real-world visual reasoning and compositional question answering},
  author={Hudson, Drew A and Manning, Christopher D},
  booktitle={Proceedings of the IEEE/CVF conference on computer vision and pattern recognition},
  pages={6700--6709},
  year={2019}
}

@inproceedings{singh2019towards,
    title={Towards VQA Models That Can Read},
    author={Singh, Amanpreet and Natarjan, Vivek and Shah, Meet and Jiang, Yu and Chen, Xinlei and Batra, Dhruv and Parikh, Devi and Rohrbach, Marcus},
    booktitle={Proceedings of the IEEE Conference on Computer Vision and Pattern Recognition},
    pages={8317-8326},
    year={2019}
}

@article{tong2024cambrian,
  title={Cambrian-1: A fully open, vision-centric exploration of multimodal llms},
  author={Tong, Shengbang and Brown, Ellis and Wu, Penghao and Woo, Sanghyun and Middepogu, Manoj and Akula, Sai C and Yang, Jihan and Yang, Shusheng and Iyer, Adithya and Pan, Xichen and others},
  journal={Advances in Neural Information Processing Systems},
  volume={37},
  pages={87310--87356},
  year={2024}
}

@inproceedings{kembhavi2016diagram,
  title={A diagram is worth a dozen images},
  author={Kembhavi, Aniruddha and Salvato, Mike and Kolve, Eric and Seo, Minjoon and Hajishirzi, Hannaneh and Farhadi, Ali},
  booktitle={European conference on computer vision},
  pages={235--251},
  year={2016},
  organization={Springer}
}

@article{geiger2021causal,
  title={Causal abstractions of neural networks},
  author={Geiger, Atticus and Lu, Hanson and Icard, Thomas and Potts, Christopher},
  journal={Advances in neural information processing systems},
  volume={34},
  pages={9574--9586},
  year={2021}
}

@inproceedings{shang2025prumerge,
  title={Llava-prumerge: Adaptive token reduction for efficient large multimodal models},
  author={Shang, Yuzhang and Cai, Mu and Xu, Bingxin and Lee, Yong Jae and Yan, Yan},
  booktitle={Proceedings of the IEEE/CVF International Conference on Computer Vision},
  pages={22857--22867},
  year={2025}
}

@article{zhang2024sparsevlm,
  title={Sparsevlm: Visual token sparsification for efficient vision-language model inference},
  author={Zhang, Yuan and Fan, Chun-Kai and Ma, Junpeng and Zheng, Wenzhao and Huang, Tao and Cheng, Kuan and Gudovskiy, Denis and Okuno, Tomoyuki and Nakata, Yohei and Keutzer, Kurt and others},
  journal={arXiv preprint arXiv:2410.04417},
  year={2024}
}

@article{heimersheim2024activation,
  title={How to use and interpret activation patching},
  author={Heimersheim, Stefan and Nanda, Neel},
  journal={arXiv preprint arXiv:2404.15255},
  year={2024}
}

@inproceedings{zhang2023activationpatching,
  title={Towards best practices of activation patching in language models: Metrics and methods},
  author={Zhang, Fred and Nanda, Neel},
  booktitle={International Conference on Learning Representations},
  volume={2024},
  pages={1651--1678},
  year={2024}
}

@article{kramar2024atp,
  title={Atp*: An efficient and scalable method for localizing llm behaviour to components},
  author={Kram{\'a}r, J{\'a}nos and Lieberum, Tom and Shah, Rohin and Nanda, Neel},
  journal={arXiv preprint arXiv:2403.00745},
  year={2024}
}

@misc{qwen2.5-VL,
    title = {Qwen2.5-VL},
    url = {https://qwenlm.github.io/blog/qwen2.5-vl/},
    author = {Qwen Team},
    month = {January},
    year = {2025}
}

@article{Qwen2VL,
  title={Qwen2-VL: Enhancing Vision-Language Model's Perception of the World at Any Resolution},
  author={Wang, Peng and Bai, Shuai and Tan, Sinan and Wang, Shijie and Fan, Zhihao and Bai, Jinze and Chen, Keqin and Liu, Xuejing and Wang, Jialin and Ge, Wenbin and Fan, Yang and Dang, Kai and Du, Mengfei and Ren, Xuancheng and Men, Rui and Liu, Dayiheng and Zhou, Chang and Zhou, Jingren and Lin, Junyang},
  journal={arXiv preprint arXiv:2409.12191},
  year={2024}
}

@article{Qwen-VL,
  title={Qwen-VL: A Versatile Vision-Language Model for Understanding, Localization, Text Reading, and Beyond},
  author={Bai, Jinze and Bai, Shuai and Yang, Shusheng and Wang, Shijie and Tan, Sinan and Wang, Peng and Lin, Junyang and Zhou, Chang and Zhou, Jingren},
  journal={arXiv preprint arXiv:2308.12966},
  year={2023}
}

@inproceedings{liu2024improved,
  title={Improved baselines with visual instruction tuning},
  author={Liu, Haotian and Li, Chunyuan and Li, Yuheng and Lee, Yong Jae},
  booktitle={Proceedings of the IEEE/CVF conference on computer vision and pattern recognition},
  pages={26296--26306},
  year={2024}
}

@misc{liu2024llavanext,
  title={Llavanext: Improved reasoning, ocr, and world knowledge},
  author={Liu, Haotian and Li, Chunyuan and Li, Yuheng and Li, Bo and Zhang, Yuanhan and Shen, Sheng and Lee, Yong Jae},
  year={2024}
}

@article{liu2023visual,
  title={Visual instruction tuning},
  author={Liu, Haotian and Li, Chunyuan and Wu, Qingyang and Lee, Yong Jae},
  journal={Advances in neural information processing systems},
  volume={36},
  pages={34892--34916},
  year={2023}
}

@inproceedings{chen2024image,
  title={An image is worth 1/2 tokens after layer 2: Plug-and-play inference acceleration for large vision-language models},
  author={Chen, Liang and Zhao, Haozhe and Liu, Tianyu and Bai, Shuai and Lin, Junyang and Zhou, Chang and Chang, Baobao},
  booktitle={European Conference on Computer Vision},
  pages={19--35},
  year={2024},
  organization={Springer}
}

@article{xing2024pyramiddrop,
  title={Pyramiddrop: Accelerating your large vision-language models via pyramid visual redundancy reduction},
  author={Xing, Long and Huang, Qidong and Dong, Xiaoyi and Lu, Jiajie and Zhang, Pan and Zang, Yuhang and Cao, Yuhang and He, Conghui and Wang, Jiaqi and Wu, Feng and others},
  journal={arXiv preprint arXiv:2410.17247},
  year={2024}
}

@inproceedings{yang2023pruning,
  title={Pruning parameterization with bi-level optimization for efficient semantic segmentation on the edge},
  author={Yang, Changdi and Zhao, Pu and Li, Yanyu and others},
  booktitle={CVPR},
  year={2023}
}

@inproceedings{rtseg,
author = {Li, Yanyu and Yang, Changdi and Zhao, Pu and others},
title = {Towards real-time segmentation on the edge},
year = {2023},
isbn = {978-1-57735-880-0},
url = {https://doi.org/10.1609/aaai.v37i2.25232},
doi = {10.1609/aaai.v37i2.25232},
articleno = {163},
numpages = {9},
series = {AAAI'23/IAAI'23/EAAI'23}
}

@article{zhang2022advancing,
  title={Advancing model pruning via bi-level optimization},
  author={Zhang, Yihua and Yao, Yuguang and Ram, Parikshit and others},
  journal={NeurIPS},
  year={2022}
}

@article{li2022pruning,
  title={Pruning-as-search: Efficient neural architecture search via channel pruning and structural reparameterization},
  author={Li, Yanyu and Zhao, Pu and Yuan, Geng and Lin, Xue and Wang, Yanzhi and Chen, Xin},
  journal={arXiv preprint arXiv:2206.01198},
  year={2022}
}

@inproceedings{zhao-etal-2024-pruning,
    title = "Pruning Foundation Models for High Accuracy without Retraining",
    author = "Zhao, Pu  and
      Sun, Fei  and
      Shen, Xuan  and
others",
    booktitle = "Findings of EMNLP 2024",
    year = "2024",
    publisher = "ACL",
    url = "https://aclanthology.org/2024.findings-emnlp.566",
    doi = "10.18653/v1/2024.findings-emnlp.566"
}

@inproceedings{zhan-etal-2024-rethinking-token,
    title = {Rethinking Token Reduction for State Space Models},
    author = {Zhan, Zheng  and Wu, Yushu  and Kong, Zhenglun  and others},
    booktitle = {EMNLP},
    month = {nov},
    year = {2024},
    address = {Miami, Florida, USA},
    publisher = {ACL},
    url = {https://aclanthology.org/2024.emnlp-main.100},
    pages = {1686--1697}
}

@inproceedings{
shen2024search,
title={Search for Efficient Large Language Models},
author={Xuan Shen and Pu Zhao and Yifan Gong and Zhenglun Kong and Zheng Zhan and Yushu Wu and Ming Lin and Chao Wu and Xue Lin and Yanzhi Wang},
booktitle={NeurIPS},
year={2024},
}

@inproceedings{
zhan2024fast,
title={Fast and Memory-Efficient Video Diffusion Using Streamlined Inference},
author={Zheng Zhan and Yushu Wu and Yifan Gong and others},
booktitle={NeurIPS},
year={2024},
url={https://openreview.net/forum?id=iNvXYQrkpi}
}

@inproceedings{
zhan2024exploring,
title={Exploring Token Pruning in Vision State Space Models},
author={Zheng Zhan and Zhenglun Kong and Yifan Gong and others},
booktitle={NeurIPS},
year={2024},
url={https://openreview.net/forum?id=eWiGn0Fcdx}
}

@INPROCEEDINGS{shen2024numerical,
  title={Numerical pruning for efficient autoregressive models},
  author={Shen, Xuan and Song, Zhao and Zhou, Yufa and others},
  booktitle={AAAI},
  year={2025}
}

@INPROCEEDINGS{shen2024lazydit,
  title={Lazydit: Lazy learning for the acceleration of diffusion transformers},
  author={Shen, Xuan and Song, Zhao and Zhou, Yufa and  others},
  booktitle={AAAI},
  year={2025}
}

@inproceedings{
shen2025sparse,
title={Sparse Learning for State Space Models on Mobile},
author={Xuan Shen and Hangyu Zheng and Yifan Gong and others},
booktitle={ICLR},
year={2025},
url={https://openreview.net/forum?id=t8KLjiFNwn}
}

@INPROCEEDINGS{shen2025quartdepth,
  title={QuartDepth: Post-Training Quantization for Real-Time Depth Estimation on the Edge},
  author={Shen, Xuan and Ma, Weize and Liu, Jing and others},
  booktitle={CVPR}, 
  year={2025}
}

@article{zhao2024fully,
  title={Fully Open Source Moxin-7B Technical Report},
  author={Zhao, Pu and Shen, Xuan and Kong, Zhenglun and Shen, Yixin and Chang, Sung-En and Rupprecht, Timothy and Lu, Lei and Nan, Enfu and Yang, Changdi and He, Yumei and others},
  journal={arXiv preprint arXiv:2412.06845},
  year={2024}
}

@article{shen2025efficient,
  title={Efficient Reasoning with Hidden Thinking},
  author={Shen, Xuan and Wang, Yizhou and others},
  journal={arXiv preprint arXiv:2501.19201},
  year={2025}
}

@inproceedings{
shen2025fastcar,
title={Fastcar: Cache Attentive Replay for Fast Auto-Regressive Video Generation on the Edge},
author={Xuan Shen and Weize Ma and Yufa Zhou   and others},
booktitle={ICLR},
year={2026}
}

@article{shen2025draftattention,
  title={DraftAttention: Fast Video Diffusion via Low-Resolution Attention Guidance},
  author={Shen, Xuan and Han, Chenxia and Zhou, Yufa and others},
  journal={arXiv preprint arXiv:2505.14708},
  year={2025}
}

@article{zhao2025taming,
  title={Taming diffusion for dataset distillation with high representativeness},
  author={Zhao, Lin and Wu, Yushu and Jiang, Xinru and Gu, Jianyang and Wang, Yanzhi and Xu, Xiaolin and Zhao, Pu and Lin, Xue},
  journal={arXiv preprint arXiv:2505.18399},
  year={2025}
}

@article{zhao2026hieramp,
  title={Hieramp: Coarse-to-fine autoregressive amplification for generative dataset distillation},
  author={Zhao, Lin and Jiang, Xinru and Xiao, Xi and Fan, Qihui and Lu, Lei and Wang, Yanzhi and Lin, Xue and Camps, Octavia and Zhao, Pu and Gu, Jianyang},
  journal={arXiv preprint arXiv:2603.06932},
  year={2026}
}

@article{zhao2026omnimem,
  title={OmniMem: Scalable and Adaptive Memory Retrieval for Long Video Generation},
  author={Zhao, Lin and Wu, Yushu and Gong, Yifan and Wang, Yanzhi and Zhao, Pu},
  journal={arXiv preprint arXiv:2605.30519},
  year={2026}
}

@article{zhao2025open,
  title={Open-Source Multimodal Moxin Models with Moxin-VLM and Moxin-VLA},
  author={Zhao, Pu and Akbari, Arash and Shen, Xuan and   others},
  journal={arXiv preprint arXiv:2512.22208},
  year={2025}
}

@article{kong2025,
      title={Token Reduction Should Go Beyond Efficiency in Generative Models -- From Vision, Language to Multimodality},
      author={Zhenglun Kong and Yize Li and Fanhu Zeng and others},
      journal={arXiv preprint arXiv:2505.18227},
      year={2025}
}

@article{yang2026survey,
  title={A Survey of Advancing Audio Super-Resolution and Bandwidth Extension from Discriminative to Generative Models},
  author={Yang, Ningyuan and Li, Yize and Cuji, Diego A and Corey, Ryan M and Zhao, Pu and Lin, Xue and Singer, Andrew C},
  journal={arXiv preprint arXiv:2605.16681},
  year={2026}
}

@article{mi2026effective,
  title={Effective MoE-based LLM Compression by Exploiting Heterogeneous Inter-Group Experts Routing Frequency and Information Density},
  author={Mi, Zhendong and Chen, Yixiao and Zhao, Pu and others},
  journal={arXiv preprint arXiv:2602.09316},
  year={2026}
}

@inproceedings{chen2024fastv,
  title={An image is worth 1/2 tokens after layer 2: Plug-and-play inference acceleration for large vision-language models},
  author={Chen, Liang and Zhao, Haozhe and Liu, Tianyu and Bai, Shuai and Lin, Junyang and Zhou, Chang and Chang, Baobao},
  booktitle={European Conference on Computer Vision},
  pages={19--35},
  year={2024},
  organization={Springer}
}
